\documentclass[conference]{IEEEtran}
\IEEEoverridecommandlockouts
\usepackage{url}
\usepackage{cite}
\usepackage{amsmath,amssymb,amsfonts}
\usepackage{algorithmic}
\usepackage{graphicx}
\usepackage{textcomp}
\usepackage{xcolor}
\usepackage{multirow}
\usepackage{tikz}
\usepackage{microtype}
\usetikzlibrary{shapes.geometric, arrows.meta, positioning, fit, backgrounds, calc}
\usetikzlibrary{arrows.meta, positioning, calc, decorations.pathreplacing}
\def\BibTeX{{\rm B\kern-.05em{\sc i\kern-.025em b}\kern-.08em
    T\kern-.1667em\lower.7ex\hbox{E}\kern-.125emX}}

\renewcommand{\baselinestretch}{0.92}

\makeatletter
\g@addto@macro\normalsize{%
  \setlength{\abovedisplayskip}{4.0pt plus 1.0pt minus 1.0pt}%
  \setlength{\belowdisplayskip}{4.0pt plus 1.0pt minus 1.0pt}%
  \setlength{\abovedisplayshortskip}{1.0pt plus 1.0pt}%
  \setlength{\belowdisplayshortskip}{1.0pt plus 1.0pt}%                  
}
\makeatother
\begin{document}

\title{An Asynchronous Two-Speed Kalman Filter for Real-Time UUV Cooperative Navigation Under Acoustic Delays\\
\thanks{This research was partially funded by Postgraduate Research Scholarship (PGRS) at Xi’an Jiaotong-Liverpool University (FOS2312JBD01), Suzhou Municipal Key Laboratory Broadband Wireless Access Technology (BWAT) and JITRI Supervision Support Fund (JSF10120220008) of XJTLU-JITRI Academy.}
\thanks{\copyright~2026 IEEE. Personal use of this material is permitted. Permission from IEEE must be obtained for all other uses, in any current or future media, including reprinting/republishing this material for advertising or promotional purposes, creating new collective works, for resale or redistribution to servers or lists, or reuse of any copyrighted component of this work in other works.}

}

\author{
    \IEEEauthorblockN{
        Shuyue Li\textsuperscript{1,2,3}, 
        Miguel López-Benítez\textsuperscript{4,5}, 
        Eng Gee Lim\textsuperscript{1}, 
        Fei Ma\textsuperscript{6}, 
        Qian Dong\textsuperscript{1}, \\
        Mengze Cao\textsuperscript{3,7}, 
        Limin Yu\textsuperscript{1,*} and 
        Xiaohui Qin\textsuperscript{3,*}
    }
    \vspace{0.3em} 
    \IEEEauthorblockA{
        \footnotesize 
        \textsuperscript{1}School of Advanced Technology, Xi'an Jiaotong-Liverpool University, Suzhou, China \\
        \textsuperscript{2}School of Engineering, University of Liverpool, Liverpool, UK \\
        \textsuperscript{3}Jiangsu JITRI Tsingunited Intelligent Control Technology Co., Ltd., Wuxi, China \\
        \textsuperscript{4}School of Computer Science and Informatics, University of Liverpool, Liverpool, UK \\
        \textsuperscript{5}ARIES Research Centre, Universidad Antonio de Nebrija, Madrid, Spain \\
        \textsuperscript{6}School of Mathematics and Physics, Xi'an Jiaotong-Liverpool University (XJTLU), Suzhou, China \\
        \textsuperscript{7}College of Mechanical and Vehicle Engineering, Hunan University, Changsha, China \\
        \textsuperscript{*}Corresponding Authors: limin.yu@xjtlu.edu.cn, qinxiaohui@tsingunited.com
    }
}

\maketitle

\begin{abstract}
In Global Navigation Satellite System (GNSS)-denied underwater environments, individual unmanned underwater vehicles (UUVs) suffer from unbounded dead-reckoning drift, making collaborative navigation (CN) crucial for accurate state estimation. However, the severe communication delay inherent in underwater acoustic channels poses serious challenges to real-time state estimation. Traditional filters, such as Extended Kalman Filters (EKFs) or Unscented Kalman Filters (UKFs), usually block the main control loop while waiting for delayed data, or effectively discard Out-of-Sequence Measurements (OOSMs), resulting in serious drift. To address this, we propose an Asynchronous Two-Speed Kalman Filter (TSKF) enhanced by a novel projection mechanism, which we term Variational History Distillation (VHD). The proposed architecture decouples the estimation process into two parallel threads: a fast-rate thread that utilizes Gaussian Process (GP) compensated dead reckoning to guarantee high-frequency real-time control, and a slow-rate thread dedicated to processing asynchronously delayed collaborative information. By introducing a Finite-Length Circular State Buffer (FLCSB), the algorithm applies delayed measurements to their corresponding historical states, and utilizes a VHD-based projection to fast-forward the correction to the current time without computationally heavy recalculations. Simulation results demonstrate that the proposed TSKF maintains a trajectory error comparable to computationally intensive batch-optimization methods under severe delays (up to 30\,s). Executing in sub-millisecond time, it significantly outperforms standard EKF/UKF. The results demonstrate an effective control, communication, and computing (3C) co-design that significantly enhances the resilience of autonomous marine automation systems.
\end{abstract}


\section{Introduction}

The deployment of Unmanned Underwater Vehicle (UUV) swarms has become a fundamental paradigm for deep-sea exploration, seabed surveying, and autonomous marine inspection \cite{b1, b2}. In such Global Navigation Satellite System (GNSS)-denied environments, individual UUVs rapidly accumulate Dead-Reckoning (DR) errors, making Collaborative Navigation (CN) essential. In standard CN frameworks, navigation vehicles equipped with high-precision sensors act as mobile reference nodes, broadcasting state information to lower-cost follower UUVs via acoustic networks \cite{b18}.

Despite its necessity, CN is profoundly hindered by the physical limitations of underwater acoustic channels. Unlike airborne electromagnetic waves, acoustic signals propagate at a sluggish nominal speed of 1500\,m/s. Compounded by variable sound speed profiles, multipath fading, and network queuing protocols, dynamic communication delays (denoted as $T$) frequently escalate to tens of seconds, alongside notable packet dropout rates \cite{b3}.

This severe latency creates a critical operational bottleneck for real-time UUV state estimation. Standard recursive Bayesian estimators, such as the Extended Kalman Filter (EKF) and Unscented Kalman Filter (UKF), are strictly bound by the Markov assumption. When confronted with severely delayed collaborative data---formally categorized as Out-of-Sequence Measurements (OOSMs)---these traditional architectures face an intractable dilemma. They must either pause the synchronous control loop to wait for delayed acoustic packets (violating real-time thruster constraints) or erroneously fuse historical data as if it perfectly represents the current UUV state. The latter induces significant model mismatches, unmanageable innovation spikes, and rapid trajectory divergence \cite{b8, b10}.

Recently, batch optimization techniques like Factor Graph Optimization (FGO) have shown mathematically robust capabilities in handling asynchronous data by iteratively re-optimizing historical trajectory windows \cite{b6, b13, b15}. However, their computational burden scales non-linearly (often super-linearly with large state augmentations) as the delay window expands. For resource-constrained UUVs powered by embedded microprocessors, these non-deterministic computational spikes make near real-time autonomous operation practically unattainable.

To bridge the critical gap between real-time execution capabilities and high-latency estimation accuracy, this paper proposes an Asynchronous Two-Speed Kalman Filter (TSKF) architecture. By integrating Gaussian Process (GP) residual kinematics \cite{b4, b7} with an asynchronous approximate Bayesian projection \cite{b11}, we engineer an efficient framework specifically designed for harsh marine deployment. 

The core contributions of this study are threefold:
\begin{enumerate}
    \item \textbf{Dual-thread architecture:} We design a decoupled filtering structure. A high-priority fast thread utilizes GP-compensated kinematics to ensure uninterrupted real-time control, while a low-priority slow thread asynchronously processes delayed acoustic measurements in the background.
    \item \textbf{Buffer-based asynchronous update:} We introduce a Finite-Length Circular State Buffer (FLCSB) coupled with a Variational History Distillation (VHD) fast-forward mechanism. This allows the filter to precisely apply delayed measurements to their true historical timestamps ($t-T$) and probabilistically project the resulting correction gradient to the current time, bypassing significant matrix re-inversions.
    \item \textbf{Rigorous evaluation via Aqua-Sim FG:} To verify our method, we evaluated the proposed TSKF within a highly realistic acoustic environment using the advanced Aqua-Sim Fourth-Generation (FG) framework on ns-3. Results demonstrate that even under severe communication delays (up to 30\,s), TSKF effectively mitigates tracking divergence, significantly outperforming standard methods. Furthermore, it achieves tracking accuracy comparable to FGO while keeping execution time strictly bounded within milliseconds for highly efficient near real-time operation.
\end{enumerate}

Ultimately, the proposed TSKF framework embodies an effective control, communication, and computing (3C) co-design paradigm: it tolerates severe acoustic communication delays, minimizes computational overhead to suit embedded constraints, and ensures uninterrupted high-frequency control.

\section{Related Work}

\subsection{Cooperative Navigation and Acoustic Latency}
CN heavily depends on integrating high-frequency proprioceptive data, such as from an Inertial Measurement Unit (IMU) and a Doppler Velocity Log (DVL), with sporadic, delayed acoustic packets. As highlighted in recent comprehensive reviews \cite{b2}, prolonged acoustic latency remains the paramount vulnerability for autonomous UUV swarms. Most conventional centralized fusion models fail severely during these delays, forcing the navigator to fall back on basic DR and endure unbounded drift \cite{b3}. While recent efforts have explored information-driven trajectory planning and motion optimization to mitigate intermittent communication \cite{b5, b19}, ensuring algorithmic resilience at the individual filter level remains the most fundamental requirement.

\subsection{Out-of-Sequence Measurements (OOSMs)}
In classical estimation literature, delayed data fusion is treated as the OOSM problem \cite{b10}. Conventional solutions, such as Bar-Shalom's algorithms, resolve this by retrodicting states or extensively augmenting the state vector to encapsulate historical trajectories (often referred to as the Augmented EKF, or Aug-EKF) \cite{b8, b10}. While mathematically elegant for millisecond-level radar tracking, directly applying these to UUV acoustic environments (where delays reach 30 seconds) triggers severe dimension explosion. The requirement to maintain thousands of augmented states causes unacceptable memory allocation and computational bottlenecks on embedded UUV hardware.

\subsection{Data-Driven and Variational Methods}
Data-driven approaches are increasingly utilized to counter model inaccuracies during signal outages \cite{b9, b12}. However, purely deep-learning models, such as Long Short-Term Memory (LSTM) networks, struggle with physical interpretability and generalization. Consequently, non-parametric GP models have gained traction due to their ability to reliably bound uncertainty while learning complex hydrodynamic residuals \cite{b4, b7}. Simultaneously, variational approximation methods have shown promise in generating virtual measurements for asynchronous updates \cite{b11}. Building upon these core concepts, we leverage the VHD mechanism—recently theorized in \cite{b16}—to compress global trajectory trends into a robust, forward-propagating correction gradient, efficiently bypassing traditional OOSM computational hurdles. In this work, the term VHD refers specifically to an approximate correction-projection mechanism, distinguishing it from full-scale variational optimization frameworks.

\section{Problem Formulation}

To implement the asynchronous historical update accurately, we establish a practical time-synchronization framework across the fleet. Specifically, cooperative UUVs are assumed to maintain strict inter-vehicle clock synchronization (e.g., typically achieved by equipping UUVs with chip-scale atomic clocks, supported by one-way travel time acoustic protocols \cite{b17}). All broadcasted acoustic packets are stamped with their precise generation time $t-T$. The residual synchronization error is assumed negligible compared with the acoustic propagation delay.

\subsection{System Kinematics and Coordinate Frames}
We consider a UUV navigating within a cooperative network, utilizing a North-East-Down (NED) navigation frame $\{N\}$ and a body frame $\{B\}$. At discrete time step $k$, the state vector $x_k \in \mathbb{R}^{n}$ encompasses the 3D position $p^n_k$, linear velocity $v^b_k$, and Euler angles $\Theta_k$:
\begin{equation}
x_k = [x, y, z, u, v, w, \phi, \theta, \psi]^{\top}.
\end{equation}
The discrete-time kinematic model, driven by high-frequency inputs $u_{k-1}$ (IMU accelerations and angular rates), is:
\begin{equation}
x_{k} = f(x_{k-1}, u_{k-1}) + w_{k-1} + f_{\text{res}}(x_{k-1}, u_{k-1}), \label{eq:motion}
\end{equation}
where $f(\cdot)$ represents the non-linear six degrees of freedom (6-DOF) kinematic transition utilizing the Jacobian matrix $J(\Theta)$, and $w_{k-1} \sim \mathcal{N}(0, Q)$ is Gaussian noise. Crucially, $f_{\text{res}}(\cdot)$ captures the highly non-linear, unmodeled hydrodynamic residual dynamics (e.g., Coriolis effects, unpredictable ocean currents) that analytical models consistently fail to estimate.

\subsection{Asynchronous Delayed Measurement Model}
The UUV periodically receives collaborative acoustic measurements $z$ from a reference vehicle. Due to propagation physics, a measurement physically arriving at the UUV's modem at current time $k$ was generated at an earlier step $k-d$. Here, $d$ denotes the delay in discrete steps, mapping to absolute time $T = d \times \Delta t$:
\begin{equation}
z_{k}^{\text{rcv}} = h(x_{k-d}) + v_{k-d} \label{eq:measurement},
\end{equation}
where $h(\cdot)$ is the observation function and $v_{k-d} \sim \mathcal{N}(0, R)$ represents acoustic noise.

The primary objective is to maintain a continuous and low-latency real-time state estimate $\hat{x}_k$ (e.g., 100\,Hz) for robust thruster control, while asynchronously incorporating severely delayed measurements $z_{k}^{\text{rcv}}$ upon arrival without violating Markov properties or stalling the processor.

\section{Proposed Asynchronous Two-Speed Filter}

To satisfy the contradictory constraints of real-time execution and delayed data integration, we propose the decoupled TSKF architecture illustrated in Fig.~\ref{fig1}.

\begin{figure}[t]
    \centering
    \resizebox{\columnwidth}{!}{%
    \begin{tikzpicture}[
        auto,
        node distance=1.0cm,
        block/.style = {rectangle, draw, thick, text centered, rounded corners, minimum height=3em, font=\footnotesize},
        fast/.style = {block, draw=blue!70, fill=blue!5, text width=7em},
        slow/.style = {block, draw=red!70, fill=red!5, text width=7em, dashed},
        buf/.style = {cylinder, draw=orange!80, fill=orange!10, shape border rotate=90, aspect=0.25, text width=5em, text centered, minimum height=3.5em, font=\scriptsize},
        sum/.style = {circle, draw, thick, minimum size=5mm, font=\small},
        line/.style = {draw, thick, -{Latex[length=2mm]}}
    ]

        \node [fast] (imu) {Proprioceptive Sensors};
        \node [fast, right=1.2cm of imu] (gp) {\textbf{Fast Thread}\\Prediction};
        \node [sum, right=0.8cm of gp] (adder) {$+$};
        \node [fast, right=0.8cm of adder, fill=green!5, draw=green!60] (ctrl) {Real-Time\\Controller};
        
        \node [buf, below=1.2cm of gp] (buffer) {Circular\\Buffer $\mathcal{B}$};
        
        \node [slow, below=1.2cm of buffer] (ekf) {\textbf{Slow Thread}\\Historical Update};
        \node [slow, right=1.2cm of ekf] (vhd) {VHD\\Projection};
        \node [slow, left=1.2cm of ekf] (modem) {Acoustic\\Modem};
       
        \draw [line] (imu) -- (gp);
        \draw [line] (gp) -- node[above, font=\tiny] {$\hat{x}^f$} (adder);
        \draw [line] (adder) -- (ctrl);

        \draw [line] (gp) -- node[right, font=\tiny] {Push} (buffer);
        \draw [line] (buffer) -- node[right, font=\tiny] {Retrieve} (ekf);
        \draw [line, dashed] (modem) -- (ekf);
        \draw [line] (ekf) -- (vhd);

        \draw [line, darkgray] (vhd.east) -- ++(0.5,0) -- ++(0, 1.8) -| (adder.south)
            node[pos=0.9, left, font=\tiny, xshift=-2pt] {Correction};
       
        \begin{scope}[on background layer]
            \draw[blue!20, dashed, thick, rounded corners] ($(imu.north west)+(-0.15,0.25)$) rectangle ($(ctrl.south east)+(0.15,-0.25)$);
            \node[blue!70, font=\tiny\bfseries] at ($(imu.north west)+(0.8,0.4)$) {Synchronous Loop};
            
            \draw[red!20, dashed, thick, rounded corners] ($(modem.north west)+(-0.15,0.25)$) rectangle ($(vhd.south east)+(0.15,-0.25)$);
            \node[red!70, font=\tiny\bfseries] at ($(modem.north west)+(0.8,0.4)$) {Asynchronous Thread};
        \end{scope}

    \end{tikzpicture}}
    \caption{Finalized TSKF architecture. The asynchronous correction path is routed externally to ensure zero overlap with inner control modules.}
    \vspace{-0.2cm}
    \label{fig1}
\end{figure}

\subsection{Fast-Rate Thread: Real-Time GP-Compensated Tracking}

The fast-rate thread operates synchronously with high-frequency onboard sensors. Its primary mandate is to furnish uninterrupted, robust state estimates to the UUV's navigation controller, agnostic to the availability of acoustic data.

To suppress rapid trajectory drift during long intervals of acoustic delay ($T$), we utilize a Sparse Gaussian Process (SGP) residual learner \cite{b4, b7}. Based on a sliding window of recent kinematic features $\mathcal{D} = \{X, Y\}$, where $X$ and $Y$ denote the training inputs and residual targets, respectively, and assuming a Squared Exponential (SE) kernel $k(x, x') = \sigma_f^2 \exp(-\frac{||x-x'||^2}{2l^2})$, the predictive mean of the residual dynamics for a query point $x_*$ is computed as:
\begin{equation}
\hat{f}_{\text{res}}(x_*) = K_{X*}^{\top}(K_{XX} + \sigma_n^2 I)^{-1}Y, \label{eq:gpmean}
\end{equation}
where $K_{XX}$ is the training input covariance matrix, $K_{X*}$ is the cross-covariance vector, $\sigma_n^2$ is the noise variance, while $l$ and $\sigma_f^2$ are the length-scale and signal variance of the SE kernel, respectively.

The fast-rate prediction step is thus fundamentally enhanced:
\begin{equation}
\hat{x}_{k|k-1} = f(\hat{x}_{k-1|k-1}, u_{k-1}) + \hat{f}_{\text{res}}(x_{k-1}) \label{eq:predict}.
\end{equation}

Simultaneously, the prediction covariance is updated using the standard Jacobian $F_k = \frac{\partial f}{\partial x}|_{\hat{x}_{k-1}}$, augmented by the GP predictive variance $\Sigma_{\text{res}} \in \mathbb{R}^{n \times n}$, which denotes the covariance induced by GP residual uncertainty:
\begin{equation}
P_{k|k-1} = F_k P_{k-1|k-1} F_k^{\top} + Q + \Sigma_{\text{res}}. \label{eq:cov}
\end{equation}

\begin{figure}[t]
  \centering
  \includegraphics[width=0.85\columnwidth]{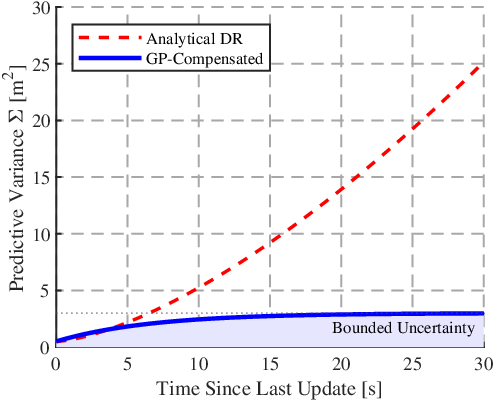}
  \caption{Predicted variance boundary of the GP residual learner during long-term acoustic interruption. Compared with uncompensated analytical position extrapolation, GP can effectively suppress error drift.}
  \label{fig2}
  \vspace{-0.2cm}
\end{figure}

Regarding buffer management, at every fast-rate step $k$, the state estimate $\hat{x}_{k|k-1}$, the covariance $P_{k|k-1}$, and the transition matrix $F_{k}$ are pushed into a FLCSB $\mathcal{B}$. The capacity is defined as $N_{\max} \ge T_{\max}/\Delta t$, ensuring the filter safely retains historical states covering the maximum expected acoustic delay. As visually confirmed in Fig.~\ref{fig2}, the prediction variance $\Sigma_{res}$ remains strictly bounded over time, effectively suppressing the boundless error drift typical of uncompensated analytical dead reckoning during communication outages.

\subsection{Slow-Rate Thread: Asynchronous VHD Update}

Functioning as an independent, event-driven background task, the slow-rate thread remains idle to conserve CPU resources. It is awakened exclusively when the modem successfully decodes collaborative measurement data $z_{k}^{\text{rcv}}$. This triggers an asynchronous interrupt, resolving the timing conflicts mapped in Fig.~\ref{fig3}.

\begin{figure}[t]
    \centering
    \resizebox{0.9\columnwidth}{!}{%
    \begin{tikzpicture}[
        timestamp/.style={circle, fill=black, inner sep=1.2pt},
        labelnode/.style={font=\scriptsize, align=center},
        arrow/.style={-{Latex[length=1.5mm, width=1mm]}, thick}
    ]
        \draw[thick, ->] (0,0) -- (7.5,0) node[right, font=\scriptsize] {Time};
        \draw[thick, ->] (0,-1.8) -- (7.5,-1.8) node[right, font=\scriptsize] {Filter};
       
        \node[timestamp] (tkd) at (1.5,0) {};
        \node[below=3pt of tkd, font=\scriptsize] {$t_{k-d}$};
        \node[above=3pt of tkd, labelnode] {Signal\\Gen.};
       
        \node[timestamp] (tk) at (5.5,0) {};
        \node[below=3pt of tk, font=\scriptsize] {$t_k$};
        \node[above=3pt of tk, labelnode] {Meas.\\Rcv.};

        \draw[<->, >=stealth, thin] (1.5,0.8) -- node[above, font=\tiny] {Delay $T$} (5.5,0.8);
     
        \node[timestamp, red] (tkd_f) at (1.5,-1.8) {};
        \draw[arrow, dashed, red] (5.5,-0.2) .. controls (4.5,-1.0) and (2.5,-1.0) .. (tkd_f.north) 
            node[pos=0.5, left, font=\tiny, xshift=-10pt] {1. Lookup};
       
        \node[timestamp, blue] (tk_f) at (5.5,-1.8) {};
        \draw[arrow, blue, ultra thick] (tkd_f) -- node[below=2pt, font=\tiny, text=black] {2. Update \& 3. VHD Projection} (tk_f);
        
        \node[below=3pt of tk_f, font=\scriptsize, blue] {$\hat{x}_k^{\text{updated}}$};
    \end{tikzpicture}}
    \caption{The timing diagram of the OOSM processing mechanism. The delay measurement is traced to the exact historical state ($t_{k-d}$) applied in the buffer, and the VHD principle is used to quickly push the generated correction forward to the current time ($t_k$) without interrupting the prediction of $t_{k+1}$.}
    \vspace{-0.2cm}
    \label{fig3}
\end{figure}

\subsubsection{Historical Retrieval}
The algorithm decodes the packet to extract its generation timestamp $t-T$ (corresponding to discrete step $k-d$). It performs an efficient $\mathcal{O}(1)$ complexity memory lookup within the FLCSB $\mathcal{B}$ to retrieve the exact historical predicted state $\hat{x}_{k-d|k-d-1}$ and its associated covariance $P_{k-d|k-d-1}$.

\subsubsection{Delayed Bayesian Update}

A standard extended Kalman update is executed strictly at the historical time step $k-d$, utilizing the measurement Jacobian $H_{k-d} = \frac{\partial h}{\partial x}|_{\hat{x}_{k-d}}$:
\begin{equation}
K_{k-d} = P_{k-d|k-d-1} H_{k-d}^{\top} (H_{k-d} P_{k-d|k-d-1} H_{k-d}^{\top} + R)^{-1},
\end{equation}
\begin{equation}
\hat{x}_{k-d|k-d}^{\text{new}} = \hat{x}_{k-d|k-d-1} + K_{k-d}(z_{k}^{\text{rcv}} - h(\hat{x}_{k-d|k-d-1})).
\end{equation}

This operation yields a delayed-state correction vector localized to that specific historical period:
\begin{equation}
\Delta x_{k-d} = \hat{x}_{k-d|k-d}^{\text{new}} - \hat{x}_{k-d|k-d}^{\text{old}}. \label{eq:correction}
\end{equation}

\subsubsection{Fast Forward Projection (VHD Logic)}
The fundamental challenge lies in propagating $\Delta x_{k-d}$ forward to the current time $k$. Recalculating the entire filter loop step-by-step from $k-d$ to $k$ intrinsically violates the near real-time control constraints.

Drawing upon the principles of approximate Bayesian inference \cite{b11} and our prior theoretical framework on VHD \cite{b16}, we deploy a projection mechanism tailored for high-latency asynchronous data. Rather than viewing $\Delta x_{k-d}$ as a mere arithmetic difference, it is treated as a probabilistically distilled gradient. Because the GP-compensated fast-rate thread has robustly captured the local trajectory shape and non-linear dynamics during the delay period, the local linearity assumption holds across the latency window, provided no extreme, unmodeled sudden maneuvers occur during this period. Thus, we project this gradient forward to the current time utilizing the accumulated product of buffered state transition matrices:
\begin{equation}
\Phi(k, k-d) \approx \prod_{i=k-d}^{k-1} F_i \label{eq:stm},
\end{equation}
where $F_i$ denotes the Jacobian evaluated along the buffered trajectory. Note that the matrix product is chronologically ordered via left-multiplication, i.e., $\prod_{i=k-d}^{k-1} F_i = F_{k-1} F_{k-2} \cdots F_{k-d}$.
\begin{equation}
\Delta x_k = \Phi(k, k-d) \Delta x_{k-d} \label{eq:forward}.
\end{equation}

The real-time state estimate $\hat{x}_k^{\text{fast}}$ maintained by the fast thread is subsequently updated asynchronously via a non-blocking additive correction:
\begin{equation}
\hat{x}_k^{\text{updated}} = \hat{x}_k^{\text{fast}} + \Delta x_k \label{eq:update}.
\end{equation}

Crucially, to maintain strict filter consistency, the predictive covariance must also incorporate the information gain obtained at $k-d$. The historical covariance correction, $\Delta P_{k-d} = -K_{k-d} H_{k-d} P_{k-d|k-d-1}$, is similarly projected forward:
\begin{equation}
P_{k}^{\text{updated}} = P_{k}^{\text{fast}} + \Phi(k, k-d) \Delta P_{k-d} \Phi(k, k-d)^{\top} \label{eq:cov_update}.
\end{equation}
The projected covariance update should be interpreted as a first-order approximate consistency-preserving correction under the local linearity assumption. This projection effectively reduces the current predictive covariance, ensuring subsequent Kalman gains remain statistically sound. This two-speed paradigm limits heavy matrix inversions to a single instance in the slow thread, keeping the main loop computationally lightweight.

Unlike classical OOSM methods that suffer dimension explosion by augmenting state vectors, VHD economically projects a dense probabilistic gradient ($\Delta x_{k-d}$) via $\Phi$. As will be demonstrated in the simulation results (Section V), standard EKF severely diverges without this projection, validating VHD's structural necessity. Unlike classical smoothing methods that explicitly reconstruct historical trajectories, the proposed VHD mechanism propagates only a compressed correction gradient, thereby avoiding iterative backward-forward optimization.

\subsection{Computational Complexity Analysis}
To substantiate the efficiency of TSKF for embedded environments, we outline the representative scaling trend and approximate practical complexity per step in Table~\ref{tab_complexity}, where $n$ is the state dimension and $d$ is the number of delayed steps. Note that the exact execution burden of FGO intrinsically depends on factor sparsity, the chosen linear solver, and the number of optimization iterations; nevertheless, its iterative scaling trend remains computationally prohibitive for real-time UUV controllers.

\begin{table}[t]
\caption{Representative Scaling Trend and Approximate Practical Complexity per Step ($d$=delay length, $n$=state dim)}
\label{tab_complexity}
\centering
\resizebox{\columnwidth}{!}{%
\begin{tabular}{|l|l|l|}
\hline
\textbf{Method} & \textbf{Representative Scaling Trend} & \textbf{Scalability ($d \gg 1$)} \\ 
\hline
Standard EKF & $\mathcal{O}(n^3)$ & Low (Diverges, see Sec. V) \\
Aug-EKF & $\mathcal{O}(d^3 n^3)$ & Out-of-Memory \\
FGO (Batch) & $\mathcal{O}(d n^3)$ (Sparse) & Iterative solver (High CPU) \\
\textbf{TSKF (Ours)} & \textbf{$\mathcal{O}(d \cdot n^3)$} & \textbf{Scalable (Linear in delay d)} \\
\hline
\end{tabular}%
}
\end{table}

Owing to its linear, non-iterative scaling, TSKF comfortably processes 30-second delays within sub-millisecond envelopes on standard microcontrollers.

\section{Simulation Results}

To comprehensively validate the proposed TSKF algorithm, we deployed a high-fidelity decoupled co-simulation platform targeting the two foremost marine operational bottlenecks: maintaining near real-time execution and seamlessly handling severe asynchronous delays.

\subsection{Simulation Setup and Evaluated Baselines}

We modeled a cooperative network comprising one reference leader and two subordinate UUVs executing a lawnmower search trajectory. Unmodeled ocean currents were deliberately introduced to replicate realistic hydrodynamic disturbances. Navigation inherently relies on a 100\,Hz tactical IMU and a DVL for continuous proprioceptive dead reckoning. 

To improve the physical realism of the Underwater Acoustic Network (UAN) and to avoid unrealistic Gaussian delay assumptions, we utilized the Aqua-Sim FG framework \cite{b14} running on ns-3. The comprehensive simulation parameters for both sensors and the acoustic network are detailed in Table~\ref{tab_sim}.

\begin{table}[t]
\caption{Simulation Parameters (Sensors and Aqua-Sim FG)}
\begin{center}
\begin{tabular}{|l|l|}
\hline
\textbf{Parameter} & \textbf{Value / Specification} \\
\hline
\multicolumn{2}{|c|}{\textit{Navigation Sensors}} \\
\hline
IMU Update Rate & $100$\,Hz \\
Accel. Random Walk (ARW) & $0.05\,\text{m/s}/\sqrt{\text{hr}}$ \\
Gyro Random Walk (GRW) & $0.01$ $\text{deg}/\sqrt{\text{hr}}$ \\
DVL Velocity Noise & $\mathcal{N}(0, 0.05^2)$\,m/s \\
\hline
\multicolumn{2}{|c|}{\textit{Aqua-Sim FG Acoustic Environment}} \\
\hline
Acoustic Frequency / Bandwidth & $25$\,kHz / $5$\,kHz \\
Transmit Power & $15$\,W \\
Nominal Sound Speed & 1500\,m/s (Thorp model) \\
Medium Access Control (MAC) protocol & Broadcast ALOHA \\
Simulated Delay Range ($T$) & $5.0$\,s to $30.0$\,s (Dynamic) \\
\hline
\end{tabular}
\label{tab_sim}
\end{center}
\end{table}

\begin{figure}[t]
  \centering
  \includegraphics[width=0.85\columnwidth]{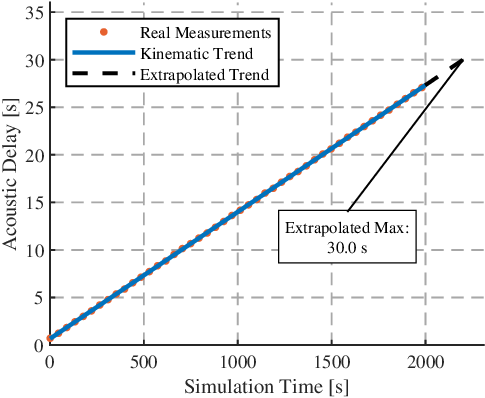}
   \caption{Dynamic acoustic propagation delay curve simulated via the Aqua-Sim FG framework, linearly extrapolated to 30\,s to evaluate the filters under extreme latency constraints (where the direct network simulation captures up to 27.3\,s).}
  \vspace{-0.2cm}
  \label{fig4}
\end{figure}

Driven by BELLHOP ray tracing and authentic MAC layer collisions (resulting in a 15\% packet loss rate), the dynamic acoustic delay successfully scales linearly from $5.0$\,s, capturing up to $27.3$\,s via direct network simulation and further projected to an extreme $30.0$\,s boundary for algorithmic stress testing, as verified in Fig.~\ref{fig4}.

We benchmarked TSKF against four distinct baselines: (1) Standard EKF (delay-ignorant); (2) Standard UKF (delay-ignorant); (3) Augmented EKF (a classic OOSM handler leveraging state augmentation \cite{b10}); and (4) FGO (the state-of-the-art for batch asynchronous fusion \cite{b6}). Tracking performance metrics were averaged across $500$ independent Monte Carlo simulations to establish statistical significance. Computational execution times were systematically profiled via MATLAB on a host machine equipped with an Intel Core i7-12700H @ 2.30 GHz processor and 16 GB of RAM, serving as a reliable relative baseline for future embedded deployments.

\subsection{Tracking Accuracy and Robustness Under Delay}

\begin{figure}[t]
  \centering
  \includegraphics[width=0.85\columnwidth]{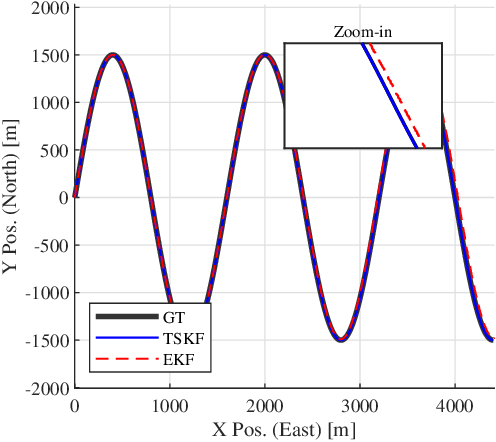}
  \caption{Simulated UUV trajectory in the $X-Y$ plane. The delay-ignorant standard EKF diverges significantly (experiencing severe delayed-update oscillations) when fusing measurements delayed by more than $10$\,s. Conversely, the proposed TSKF maintains robust adherence to the ground truth trajectory. The zoom-in inset highlights the tracking performance difference at the maximum delay boundary ($30$\,s).}
  \vspace{-0.2cm}
  \label{fig5}
\end{figure}

Because the UUV physically translates tens of meters during a severe delay $T$, forcing a measurement originating from $t-T$ into the filter as if it were a current observation fundamentally violates state space geometry. This artificially ``pulls'' the estimate backward along its trajectory, triggering the severe oscillatory divergence observed in Standard EKF/UKF.

\begin{figure}[t]
  \centering
  \includegraphics[width=0.85\columnwidth]{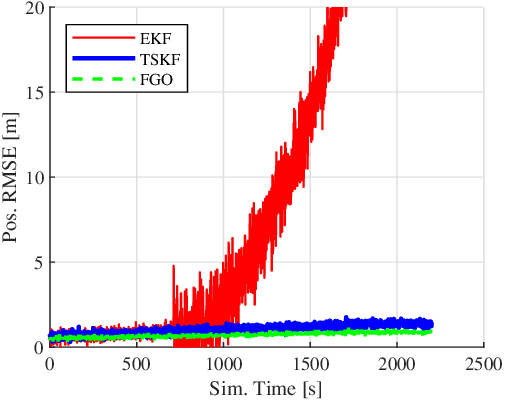}
  \caption{Position RMSE comparison under dynamic acoustic delays (evaluated up to the 30\,s extreme boundary). The proposed TSKF limits error drift during long delay periods and performs closely to the FGO benchmark.}
  \vspace{-0.2cm}
  \label{fig6}
\end{figure}

As quantitatively summarized in Table~\ref{tab_results} and explicitly illustrated in both the trajectory plot (Fig.~\ref{fig5}) and the temporal error evolution (Fig.~\ref{fig6}), standard filters exhibit severe degradation under high latency. While the standard EKF exhibits reasonable accuracy at a 10\,s delay (Root Mean Square Error, RMSE $1.06$\,m), it suffers significant drift at 20\,s (RMSE $15.49$\,m) and effectively diverges ($>$50.0\,m) at 30\,s. The augmented EKF handles 10\,s delays adequately (RMSE $1.15$\,m), but suffers from impractical memory growth at 30\,s delays owing to the unmanageable dimensionality of a covariance matrix encompassing 3000 discrete historical states. Since the standard UKF displays tracking degradation patterns nearly identical to the EKF under high-latency conditions, its metrics are omitted for graphical clarity.

As firmly demonstrated by the RMSE profiles (Fig.~\ref{fig6}), the proposed TSKF consistently suppresses error accumulation throughout prolonged delay windows. It trails the near-optimal batch FGO baseline exceptionally closely (RMSE $1.92$\,m versus $0.94$\,m at a 30\,s delay), effectively mitigating the structural damage traditionally inflicted by severe acoustic latency.

\begin{table}[t]
\caption{Tracking Accuracy and Execution Time Under Realistic Acoustic Delays}
\label{tab_results}
\centering
\begin{tabular}{|c|c|c|c|}
\hline
\multirow{2}{*}{\textbf{Algorithm}} & \textbf{Delay $T$} & \textbf{Pos. RMSE} & \textbf{Exec. Time} \\
 & \textbf{(s)} & \textbf{(m)} & \textbf{/Step ($\mu$s)} \\
\hline
Standard EKF & 10 & 1.06 & 4 \\
Aug-EKF (OOSM) & 10 & 1.15 & 38 \\
FGO (Batch) & 10 & 0.87 & 400 \\
\textbf{Proposed TSKF} & 10 & 1.27 & 33 \\
\hline
Standard EKF & 20 & 15.49 & 4 \\
Aug-EKF (OOSM) & 20 & 2.14 & 161 \\
FGO (Batch) & 20 & 0.96 & 1487 \\
\textbf{Proposed TSKF} & 20 & 1.61 & 23 \\
\hline
Standard EKF & 30 & $>$ 50.0 (Div.) & 4 \\
Aug-EKF (OOSM) & 30 & Memory-intensive & -- \\
FGO (Batch) & 30 & 0.94 & 3457 \\
\textbf{Proposed TSKF} & 30 & 1.92 & 27 \\
\hline
\end{tabular}
\end{table}
\subsection{Real-Time Computational Efficiency}

The true operational viability of a UUV algorithm hinges on its deployability on resource-constrained ARM Cortex embedded microprocessors. While FGO guarantees superior tracking accuracy, it exhibits an unfavorable non-linear growth pattern as the temporal optimization window expands. As shown in Table~\ref{tab_results}, FGO requires approximately 3.46~ms at a 30\,s delay (nearly an order of magnitude slower than at 10\,s, and over 100$\times$ slower than the proposed TSKF). This super-linear computational growth poses severe bottleneck hazards for embedded controllers.

Conversely, the proposed TSKF completes the asynchronous fusion cycle on the order of tens of microseconds (e.g., $\approx 30\,\mu$s) per step under MATLAB profiling on an Intel Core i7-12700H @ 2.30 GHz processor in the full 9-dimensional state framework, establishing an execution time curve that remains remarkably flat irrespective of the delay magnitude. These measurements primarily serve as relative comparative indicators rather than hardware-certified benchmarks. The algorithm achieves FGO-tier accuracy while running significantly faster even in the interpreted MATLAB environment, effectively mitigating the severe memory bottlenecks and scale vulnerabilities observed in Aug-EKF. This computational efficiency is expected to translate into substantial performance gains on low-power embedded microcontrollers, where non-linear iterative optimization becomes prohibitive.

From a real-world hardware deployment perspective, the TSKF architecture is highly modular for microcontrollers. The fast thread operates seamlessly within the real-time operating system inner loop, while the slow thread executes solely as a Universal Asynchronous Receiver-Transmitter (UART)-triggered background interrupt task. This ensures it consumes negligible RAM allocation ($<$50\,KB for a 30\,s delay at 100\,Hz) and practically zero CPU cycles during periods of acoustic silence.

\section{Conclusion}

This paper has introduced an Asynchronous TSKF structurally engineered to resolve the critical bottleneck of severe acoustic communication delays in UUV cooperative navigation. By decoupling the estimation process into a high-rate GP-compensated prediction thread and an event-driven, low-rate asynchronous update thread equipped with a FLCSB, the proposed architecture cleanly circumvents the OOSM blocking problem. Through probabilistically motivated VHD projections, highly delayed measurements are fused efficiently without triggering significant matrix recalculations. High-fidelity simulations via the Aqua-Sim FG framework have validated that TSKF effectively mitigates tracking divergence in the evaluated scenarios with delays up to $30$\,s. Crucially, by harmonizing communication latency tolerance, computing efficiency, and control continuity, the algorithm demonstrates a robust 3C co-design. It provides trajectory tracking accuracy comparable to complex FGO methodologies while maintaining sub-millisecond, highly scalable computational efficiency. Future work will focus on investigating robustness under conditions of imperfect clock synchronization and timestamp uncertainty, extending the applicability of TSKF to less specialized underwater hardware. This presents a highly robust and practical real-time information fusion paradigm for the next generation of autonomous ocean clusters.

\begingroup
\renewcommand{\baselinestretch}{1.0}

\endgroup

\small
\begin{thebibliography}{00}


\bibitem{b1} L. Paull, S. Saeedi, M. Seto, and H. Li, ``AUV Navigation and Localization: A Review,'' \emph{IEEE J. Ocean. Eng.}, vol. 39, no. 1, pp. 131--149, 2014.

\bibitem{b2} S. Li \emph{et al.}, ``Enabling Cooperative Autonomy in UUV Clusters: A Survey of Robust State Estimation and Information Fusion Techniques,'' \emph{Drones}, vol. 9, no. 11, art. no. 752, 2025.

\bibitem{b18} S. Choi, Y. Choi, and J. Jung, ``Improved Localization of Unmanned Underwater Vehicle via Cooperative Navigation of Unmanned Surface Vehicle Equipped with Ultrashort Baseline,'' \emph{J. Sens. Sci. Technol.}, vol. 33, no. 5, pp. 391--398, 2024.

\bibitem{b3} G. Wang, J. Zhu, S. Hu, Z. Peng, and J.-H. Cui, ``An Asynchronous Multicluster Network System for AUV Swarm Communication and Positioning: Design and Trial,'' \emph{IEEE Internet Things J.}, vol. 12, no. 13, pp. 25393--25406, 2025.

\bibitem{b8} J. Li, J. Song, and J. Liu, ``OOSM-Adaptive Sequential Tobit Kalman Filter (OS-TKF) for Distributed X-Ray Pulsar-Based Navigation Systems,'' \emph{IEEE Trans. Aerosp. Electron. Syst.}, vol. 61, no. 3, pp. 6564--6575, 2025.

\bibitem{b10} Y. Bar-Shalom, H. Chen, and M. Mallick, ``One-step solution for the multistep out-of-sequence-measurement problem in tracking,'' \emph{IEEE Trans. Aerosp. Electron. Syst.}, vol. 40, no. 1, pp. 27--37, 2004.

\bibitem{b6} Q. Li, H. Huang, Y. Sun, and Y. Ben, ``A Factor Graph With Mercator Projection-Based Cooperative Localization Algorithm for Multiple AUVs,'' \emph{IEEE Trans. Instrum. Meas.}, vol. 74, pp. 1--10, 2025.

\bibitem{b13} S. Cheng, H. Zhu, X. Qu, and Y. Wang, ``A Factor Graph-Based Multiple Sensors Fusion Method for Enhanced Underwater Navigation of Remotely Operated Vehicles,'' in \emph{Proc. IEEE ICUS}, 2024, pp. 1980--1984.

\bibitem{b15} C. Cadena \emph{et al.}, ``Past, Present, and Future of Simultaneous Localization and Mapping: Toward the Robust-Perception Age,'' \emph{IEEE Trans. Robot.}, vol. 32, no. 6, pp. 1309--1332, 2016.

\bibitem{b4} Y. Dang, Y. Huang, X. Shen, D. Zhu, and Z. Chu, ``Incremental Sparse Gaussian Process-Based Model Predictive Control for Trajectory Tracking of Unmanned Underwater Vehicles,'' \emph{IEEE Robot. Autom. Lett.}, vol. 10, no. 3, pp. 2327--2334, 2025.

\bibitem{b7} N. Cohen and I. Klein, ``Gaussian Process Regression for Improved Underwater Navigation,'' in \emph{Proc. IEEE/ION Position, Location and Navigation Symp. (PLANS)}, 2025, pp. 1125--1132.

\bibitem{b11} M. Greiff and K. Berntorp, ``Asynchronous Variational-Bayes Kalman Filtering,'' in \emph{Proc. IEEE CDC}, 2024, pp. 1987--1992.

\bibitem{b5} Z. Zeng \emph{et al.}, ``Information-driven Path Planning for Hybrid Aerial Underwater Vehicles,'' \emph{IEEE J. Ocean. Eng.}, vol. 48, no. 3, pp. 689--715, 2023.

\bibitem{b19} A. Tiranti \emph{et al.}, ``Motion Optimization Strategy for Passive Acoustic Monitoring With a Team of AUVs Considering Intermittent Communication,'' \emph{IEEE J. Ocean. Eng.}, vol. 50, no. 4, pp. 2782--2796, 2025.

\bibitem{b9} Y. Hou, L. Cheng, Q. Wang, J. Li, and Y. Ke, ``An Underwater Multisource Fusion Anomaly Detection Navigation Algorithm Based on Factor Graph and LSTM,'' \emph{IEEE Trans. Instrum. Meas.}, vol. 74, pp. 1--11, 2025.

\bibitem{b12} X. Wang \emph{et al.}, ``Underwater Federated Learning: Empowering Autonomous Underwater Vehicle Swarm with Online Learning Capabilities,'' in \emph{Proc. IEEE GLOBECOM}, 2024, pp. 379--384.

\bibitem{b16} S. Li \emph{et al.}, ``Communication Outage-Resistant UUV State Estimation: A Variational History Distillation Approach,'' \emph{arXiv preprint arXiv:2603.29512}, 2026.

\bibitem{b17} K. G. Kebkal \emph{et al.}, ``Underwater Acoustic Modems with Synchronous Chip-Scale Atomic Clocks for Scalable Tasks of AUV Underwater Positioning,'' \emph{Gyroscopy Navig.}, vol. 10, no. 4, pp. 313--321, 2019.

\bibitem{b14} J. Guo, S. Song, H. Chen, B. Huangfu, J. Liu, and J.-H. Cui, ``Aqua-Sim Fourth Generation: Toward General and Intelligent Simulation for Underwater Acoustic Networks,'' \emph{IEEE Internet Things J.}, vol. 12, no. 15, pp. 30203--30214, 2025.

\end{thebibliography}
\end{document}